\documentclass[twoside,11pt]{article}

\usepackage{jmlr2e}
\usepackage{graphicx}
\usepackage{float}
\usepackage{booktabs}

\usepackage{listings}
\usepackage{color}

\definecolor{dkgreen}{rgb}{0,0.6,0}
\definecolor{gray}{rgb}{0.5,0.5,0.5}
\definecolor{mauve}{rgb}{0.58,0,0.82}

\graphicspath{{fig/}}

\ShortHeadings{modAL: A modular active learning framework for Python}{}
\firstpageno{1}

\begin{document}

\title{modAL: A modular active learning framework for Python}

\author{\name Tivadar Danka \email danka.tivadar@brc.mta.hu \\
       \addr Biological Research Centre\\
       Hungarian Academy of Sciences\\
       Szeged, H6720, Hungary
       \AND
       \name Peter Horvath \email horvath.peter@brc.mta.hu \\
       \addr Biological Research Centre\\
	   Hungarian Academy of Sciences\\
	   Szeged, H6720, Hungary}
	   
\maketitle

\begin{abstract}
modAL is a modular active learning framework for Python, aimed to make active learning research and practice simpler. Its distinguishing features are (i) clear and modular object oriented design (ii) full compatibility with scikit-learn models and workflows. These features make fast prototyping and easy extensibility possible, aiding the development of real-life active learning pipelines and novel algorithms as well. modAL is fully open source, hosted on GitHub.\hspace{1pt}\footnote{https://github.com/modAL-python/modAL} To assure code quality, extensive unit tests are provided and continuous integration is applied. In addition, a detailed documentation with several tutorials are also available for ease of use. The framework is available in PyPI and distributed under the MIT license.
\end{abstract}


\section{Introduction}
Upon learning patterns from data in real-life applications, labelling examples often consume significant time and money, which makes it infeasible to obtain large training sets. For example, sentiment analysis of texts requires extensive manual annotation, which costs expert time. Another example is the optimization of black box functions, for which the evaluation is costly or derivatives are not available. In these cases, active learning can be used to query labels for the most informative instances. modAL is an active learning framework for Python, designed with modularity, flexibility and extensibility in mind. Built on top of scikit-learn (\cite{scikit-learn, sklearn_api}), it allows the rapid prototyping of active learning workflows with a large degree of freedom. It was designed to be easily extensible, allowing researchers to implement and test novel active learning strategies with minimal effort.

\section{Design principles and features}
Our objective with modAL was to create an active learning library which takes advantage of the advanced features of Python and the extensive ecosystem of scikit-learn, making the implementation of complex workflows simple and intuitive. Specifically, modAL was designed with the following goals in mind. \\
\begin{enumerate}
	\item \textbf{Modularity: separating and recombining parts of a workflow.} In general, an active learning workflow consists of a learning algorithm and a query strategy. In modAL, this is represented by the \verb|ActiveLearner| class, for which these components are passed upon object creation. Learning algorithms can be used with query strategies in any combination, making rapid prototyping possible.	
	\item \textbf{Extensibility: simple customization of parts.} In a modAL active learning workflow, a query strategy is simply a function, given to the object representing the active learning algorithm upon initialization. Implementing custom query strategies can be done without understanding class structures or modAL internals. Thus it requires minimal effort, allowing researchers to easily test novel strategies and compare them with existing ones. 
	\item \textbf{Flexibility: compatibility with the scikit-learn ecosystem.} scikit-learn is one of the most popular machine learning tools in Python, used by researchers and practicioners as well. modAL is built on top of it, allowing the use any of its classifier and regressor algorithms in active learning pipelines. Objects in modAL also follow the scikit-learn API, making it possible to insert them into already existing workflows.
\end{enumerate}

modAL supports a wide range of active learning algorithms for both pool-based and stream-based (\cite{stream-based-sampling}) scenarios. For multiclass problems, uncertainty sampling methods such as least confident (\cite{uncertainty-sampling}), max margin and max entropy sampling; committee-based methods such as query by committee (\cite{query-by-committee}) and query by disagreement (\cite{query-by-disagreement}); ranked batch-mode sampling (\cite{ranked-batch-mode}); expected error reduction (\cite{expected-error}) is provided. For multilabel classification, SVM binary minimum (\cite{svm-binary-minimum}); max loss and mean max loss (\cite{svm-max-loss}); MinConfidence, AvgConfidence, MinScore, AvgScore (\cite{esuli-sebastiani}) algorithms are implemented. For density weighting, the information density framework (\cite{information-density}) is available. In addition to classification, active regression. Moreover, Bayesian optimization is available with probability of improvement, expected improvement and upper confidence bound strategies (\cite{7352306}). 

\section{Classes and interfaces}\label{section:classes-and-interfaces}

For modularity and easy extensibility, active learning workflows are abstracted and represented by the \verb|ActiveLearner|, \verb|BayesianOptimizer|, \verb|Committee| and \verb|CommitteeRegressor| classes. All classes inherit from the \verb|sklearn.base.BaseEstimator| class. \verb|ActiveLearner| serves as an abstract model for general active learning algorithms, while \verb|Committee| and \verb|CommitteeRegressor| implements committee-based strategies. Bayesian optimization algorithms are represented by \verb|BayesianOptimizer|. All classes require a learner and a query strategy upon initialization. In the case of \verb|ActiveLearner| and \verb|BayesianOptimizer|, the learner is an arbitrary object implementing the scikit-learn API, while the for \verb|Committee| and \verb|CommitteeRegressor|, a list of \verb|ActiveLearner| instances must be provided. Again, the query function can be factored into two functions: one calculating the utility for each instance and one selecting the instances to be queried based upon the utility score. This modular design makes easy extensibility and interaction with other libraries possible. The use of ActiveLearner is demonstrated below. \\

\begin{lstlisting}
from modAL.models import ActiveLearner
from modAL.uncertainty import uncertainty_sampling
from sklearn.ensemble import RandomForestClassifier

# initializing the learner
learner = ActiveLearner(
    estimator=RandomForestClassifier(),
    query_strategy=uncertainty_sampling
)

# training
learner.fit(X_training, y_training)

# query for labels
query_idx, query_inst = learner.query(X_pool)
# ...obtaining new labels from the Oracle...
# supply label for queried instance
learner.teach(X_pool[query_idx], y_new)
\end{lstlisting}


\section{Comparison with other libraries}
To assess the features of modAL, a comparison between libraries is provided. We compare modAL to the Python libraries acton\hspace{1pt}\footnote{https://github.com/chengsoonong/acton}, alp\hspace{1pt}\footnote{https://github.com/davefernig/alp}, libact\hspace{1pt}\footnote{https://github.com/ntucllab/libact} (\cite{YY2017}) and the Java library JCLAL (\cite{JCLAL}) in Tables \ref{tab:algorithms_full}, \ref{tab:other-features}. The comparison is made with respect to supported algorithms, design and support. For Python libraries, a runtime comparison for least confident sampling, query by committee\footnote{A minor bugfix was applied on \textit{alp} for QBC to work, see https://github.com/davefernig/alp/issues/1} and expected error reduction can be found in Table \ref{tab:runtime}. The runtime data was obtained by averaging the result of 10 runs for each algorithm. During each run, 10 queries were made. The comparison script is available at https://github.com/modAL-python/modAL/blob/master/examples/runtime\_comparison.py.

\section{Availability}
The framework is fully open-source, hosted on GitHub.\hspace{1pt}\footnote{https://github.com/modAL-python/modAL} Besides the core features, detailed documentation and a wealth of examples and tutorials are available at the project website\hspace{1pt}\footnote{https://modAL-python.github.io}, making active learning accessible for a wide range of users. All tutorials and examples on the official website can be downloaded as a Jupyter notebook. To assure code quality, extensive unit tests are provided with $ 98\% $ code coverage. Continuous integration is applied using Travis-CI. modAL is also available from PyPI.

\acks{T.D. and P.H. acknowledges support from the European Regional Development Funds (GINOP-2.3.2-15-2016-00001, GINOP-2.3.2-15-2016-00037).}

\begin{table}\centering
	\begin{tabular}{r c c c c c c c c c c c c c c c} \toprule
		& \rotatebox[origin=c]{90}{pool} 
		& \rotatebox[origin=c]{90}{stream} 
		& \rotatebox[origin=c]{90}{regression} 
		& \rotatebox[origin=c]{90}{committee} 
		& \rotatebox[origin=c]{90}{multilabel} 
		& \rotatebox[origin=c]{90}{information} \rotatebox[origin=c]{90}{density} 
		& \rotatebox[origin=c]{90}{expected} \rotatebox[origin=c]{90}{error} 
		& \rotatebox[origin=c]{90}{variance} \rotatebox[origin=c]{90}{reduction} 
		& \rotatebox[origin=c]{90}{hierarchical} \rotatebox[origin=c]{90}{sampling} 
		& \rotatebox[origin=c]{90}{ranked batch} 
		& \rotatebox[origin=c]{90}{meta-} \rotatebox[origin=c]{90}{learning} 
		& \rotatebox[origin=c]{90}{multilabel} 
		& \rotatebox[origin=c]{90}{Bayesian} \rotatebox[origin=c]{90}{optimization}
		& \rotatebox[origin=c]{90}{cost} \rotatebox[origin=c]{90}{sensitive} \\ \midrule
		modAL & \checkmark & \checkmark & \checkmark & \checkmark & \checkmark & \checkmark & \checkmark & X & X & \checkmark & X & \checkmark & \checkmark & X\\
		acton & \checkmark & X & \checkmark & \checkmark & X & X & X & X & X & X & X & X & X & X \\
		alp & \checkmark & X & X & \checkmark & X & X & X & X & X & X & X & X & X & X \\
		libact & \checkmark & X & X & \checkmark & \checkmark & X & \checkmark & \checkmark & \checkmark & X & \checkmark & \checkmark & X & \checkmark \\
		JCLAL & \checkmark & \checkmark & X & \checkmark & \checkmark & \checkmark & \checkmark & \checkmark & X & X & X & \checkmark & X & X \\ \bottomrule
	\end{tabular}
	\caption{Comparison of libraries with respect to supported algorithms.}
	\label{tab:algorithms_full}
	
	\vspace{2em}
	
%
	
	\begin{tabular}{r c c c c c} \toprule
		& \vtop{\hbox{\strut sklearn}\hbox{\strut model usability}} & \vtop{\hbox{\strut follows}\hbox{\strut sklearn API}} & \vtop{\hbox{\strut actively}\hbox{\strut maintained}} & \vtop{\hbox{\strut Python}\hbox{\strut version}} & \vtop{\hbox{\strut documentation,}\hbox{\strut tutorials}}\\ \midrule
		modAL & \checkmark & \checkmark & \checkmark & 3 & \checkmark \\ 
		acton & with adapters & \checkmark & \checkmark & 3 & \checkmark \\
		alp & \checkmark & X & X & 2, 3 & X \\
		libact & with adapters & X & \checkmark & 2, 3 & \checkmark \\
		JCLAL & -- & -- & \checkmark & -- & \checkmark \\
		\bottomrule
	\end{tabular}
	\caption{Comparison of libraries with respect to design and support.}
	\label{tab:other-features}
	
	\vspace{2em}
	
	\begin{tabular}{r c c c} \toprule
		& least confident & QBC & EER \\ \midrule
		modAL & 0.0087 s & 0.0465 s & 2.1255 s \\
		acton & 0.1860 s & 0.5858 s & -- \\
		alp & 0.0055 s & 0.0573 s & -- \\
		libact & 0.0191 s & 0.0324 s & 2.8436 s \\ \bottomrule
	\end{tabular}
	\caption{Comparison of Python libraries with respect to runtime. For each algorithm, 10 queries were made in a run and each run was repeated 10 times.}
	\label{tab:runtime}
\end{table}


\vskip 0.2in
\bibliography{modAL}

\end{document}